%% file: acl_latex.tex
\documentclass[11pt]{article}

\usepackage[final]{acl}

\usepackage{times}
\usepackage{latexsym}
\usepackage{amsfonts}
\usepackage{amsmath}
\usepackage{amssymb}
\usepackage{CJKutf8}
\usepackage{array}
\usepackage{adjustbox}
\usepackage[T1]{fontenc}
\usepackage[table]{xcolor}
\usepackage{float}

\usepackage[utf8]{inputenc}

\usepackage{microtype}

\usepackage{inconsolata}
\usepackage{booktabs}
\usepackage{graphicx}
\usepackage{multirow}
\usepackage{makecell}

\title{\adjustbox{height=1.7em, valign=m}{\includegraphics[width=1.7em]{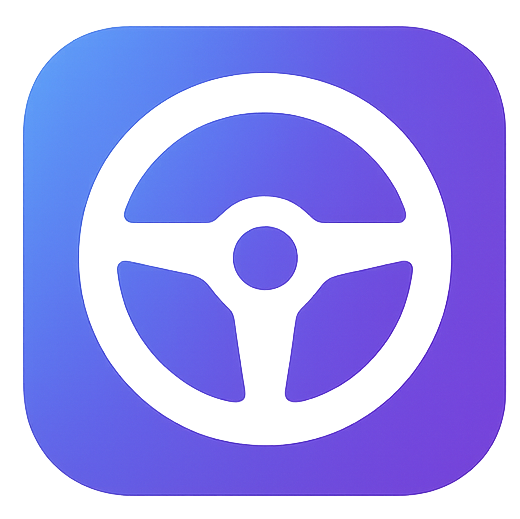}}
EasySteer: A Unified Framework for High-Performance and Extensible LLM Steering}

\author{
Haolei Xu\textsuperscript{1},
Xinyu Mei\textsuperscript{1},
Yuchen Yan\textsuperscript{1},
Rui Zhou\textsuperscript{1},
\textbf{Wenqi Zhang}\textsuperscript{1},\\
\textbf{Weiming Lu}\textsuperscript{1}\thanks{~Corresponding author.},
\textbf{Yueting Zhuang}\textsuperscript{1}, \textbf{Yongliang Shen}\textsuperscript{1}\footnotemark[1]
\\
\textsuperscript{1}Zhejiang University\\
\texttt{\{xuhaolei,luwm,syl\}@zju.edu.cn}\\
Open-source repository: \url{https://github.com/ZJU-REAL/EasySteer} \\
Demonstration video: \url{https://www.youtube.com/watch?v=3rRGzZmhrXg}
}

\begin{document}
\maketitle
\begin{abstract}
Large language model (LLM) steering has emerged as a promising paradigm for controlling model behavior at inference time through targeted manipulation of hidden states, offering a lightweight alternative to expensive retraining. However, existing steering frameworks suffer from critical limitations: computational inefficiency, limited extensibility, and restricted functionality that hinder both research progress and practical deployment.
We present EasySteer, a unified framework for high-performance, extensible LLM steering built on vLLM. Our system features modular architecture with pluggable interfaces for both analysis-based and learning-based methods, fine-grained parameter control, pre-computed steering vectors for eight application domains, and an interactive demonstration system. Through deep integration with vLLM's optimized inference engine, EasySteer achieves 10.8-22.3$\times$ speedup over existing frameworks. Extensive experiments demonstrate its effectiveness in overthinking mitigation, hallucination reduction, and other key applications.
EasySteer transforms steering from research technique to production-ready capability, establishing critical infrastructure for deployable, controllable language models.
\end{abstract}

\section{Introduction}

Large language models (LLMs) have achieved remarkable capabilities, yet controlling their behavior during deployment remains a fundamental challenge~\citep{zhao2023survey}. Fine-tuning requires expensive retraining and risks catastrophic forgetting, while prompt engineering offers only superficial control without behavioral guarantees~\citep{yao2024survey}. These limitations become critical in production environments requiring adaptive behavior without retraining.

LLM steering offers a compelling solution through targeted manipulation of hidden states during inference~\citep{turner2023steering,zhang2026locate}. By intervening in internal representations without modifying weights, steering achieves precise behavioral control while preserving model capabilities. This approach leverages the Linear Representation Hypothesis~\citep{park2023linear}, which posits that concepts are encoded as linear structures amenable to vector operations.

Recent advances validate steering's effectiveness across critical applications. Thinking pattern vectors successfully mitigate overthinking in mathematical reasoning~\citep{lin2025controlling,liu2025fractional,huang2026sat}, preference-based methods achieve personality control~\citep{cao2024personalized}, and simple additive vectors manipulate refusal behaviors~\citep{lee2024programming}. These successes establish steering as both a practical control mechanism and a tool for mechanistic interpretability~\citep{zhang2026evaluating,li2026dynamics}.

Despite these advances, practical implementation remains challenging, as steering typically requires modifying the forward propagation process through complex wrappers or hooks~\citep{ko2026vllm,blum2026xmix}, creating significant engineering barriers. Several frameworks have emerged to facilitate steering research~\citep{miehling2026ai}, including repeng~\citep{vogel2024repeng}, pyreft~\citep{wu2024reft}, and EasyEdit2~\citep{xu2025easyedit2}. However, existing steering frameworks suffer from three critical limitations (Table \ref{tab:intro}): (1) \textbf{computational inefficiency} with severe inference bottlenecks, where EasyEdit2 lacks batch inference support; (2) \textbf{lack of essential capabilities} like token-specific interventions and multi-vector coordination, limiting applicability to complex scenarios requiring conditional activation or multi-objective optimization; (3) \textbf{inflexible architectures} preventing convenient custom algorithm integration.

\input{table/table1}

To address these challenges, we present EasySteer, an Apache-2.0 licensed open-source unified framework for high-performance, extensible LLM steering built on vLLM~\citep{kwon2023efficient}. Our system comprises four integrated modules: \textbf{(1) Steering Vector Generation Module} supporting both analysis-based and learning-based methods; \textbf{(2) Steering Vector Application Module} leveraging vLLM's optimized engine for efficient hidden state intervention with pluggable algorithm interfaces and fine-grained parameter control; \textbf{(3) Comprehensive Resource Library} offering production-ready steering vectors and examples for eight application domains with documented evaluation results; \textbf{(4) Interactive Demonstration System} providing an intuitive web interface for vector extraction, training, inference and chat.

Through deep vLLM integration, EasySteer achieves 10.8-22.3× speedup over existing frameworks while maintaining 81-91\% of baseline throughput even under multi-vector configurations. Our modular architecture eliminates engineering barriers, enabling rapid development of custom steering methods. Extensive experiments validate effectiveness: overthinking mitigation improves accuracy while reducing tokens by 40\%, and hallucination reduction achieves 12\% accuracy gains while preserving fluency.
\section{Related Work}
\paragraph{Model Control Paradigms.}
Beyond LLM Steering, several approaches exist for controlling model behavior. Prompt Engineering~\citep{sahoo2024systematic} guides generation through carefully designed instruction templates and contextual framing, with Retrieval-Augmented Generation (RAG)~\citep{gao2023retrieval} extending this via dynamic knowledge integration. Fine-tuning methods adapt model behavior through weight updates, ranging from full parameter updates to Parameter-Efficient Fine-Tuning (PEFT)~\citep{han2024parameter} approaches like LoRA~\citep{hu2022lora}. Model Editing techniques~\citep{zhang2024comprehensive,zhang2026towards}, including ROME~\citep{meng2022locating} and MEMIT~\citep{meng2022mass}, enable precise knowledge updates by targeting specific parameters encoding factual information. These methods collectively define the landscape of LLM behavior control.
\paragraph{Mechanistic Interpretability.}
Research has established that neural network activations encode semantically meaningful features~\citep{mikolov2013linguistic,elhage2022toy}. Central to this understanding is the Linear Representation Hypothesis~\citep{park2023linear}, which posits that LLMs encode high-level concepts as linear structures in representation space. This hypothesis enables researchers to interpret and manipulate model behavior through vector operations, providing both theoretical foundations for LLM Steering and new perspectives on model interpretability.
\section{Formalization of LLM Steering}

Consider an L-layer language model $M$ processing input sequence $x = (x_1, x_2, \ldots, x_n)$. Let $h_{l,i} \in \mathbb{R}^d$ denote the hidden state at layer $l \in \{1, 2, \ldots, L\}$ and position $i$, where $d$ represents the hidden dimension and $n$ denotes the sequence length. We formalize LLM steering as an inference-time transformation function $f$. When a specific condition $C$ is satisfied, this function maps $h_{l,i}$ to a steered representation $h'_{l,i} = f(h_{l,i})$. The condition $C$ is determined by contextual factors or internal model states and triggers the directed representation update, thereby modulating generation behavior without modifying model weights $W$. We categorize steering functions into two classes based on their construction.

\subsection{Analysis-based Steering}
\label{sec:analysis}
Analysis-based steering comprises two phases: concept extraction and steering intervention. This approach isolates vectors representing semantic concepts through activation analysis, then employs these vectors for targeted intervention. Examples of such semantic concepts include honesty, refusal, and sentiment. Common extraction methods (e.g., CAA) are described in detail in Appendix \ref{appendix:extraction}. Given concept vector $v$, the steering function becomes:
\begin{equation}
f(h_{l,i}) := h_{l,i} + \alpha \cdot v
\end{equation}
where $\alpha \in \mathbb{R}$ controls steering intensity and direction. Positive values of $\alpha$ enhance the concept, while negative values suppress it.
\begin{figure*}[htbp]
    \centering
    \includegraphics[width=\textwidth]{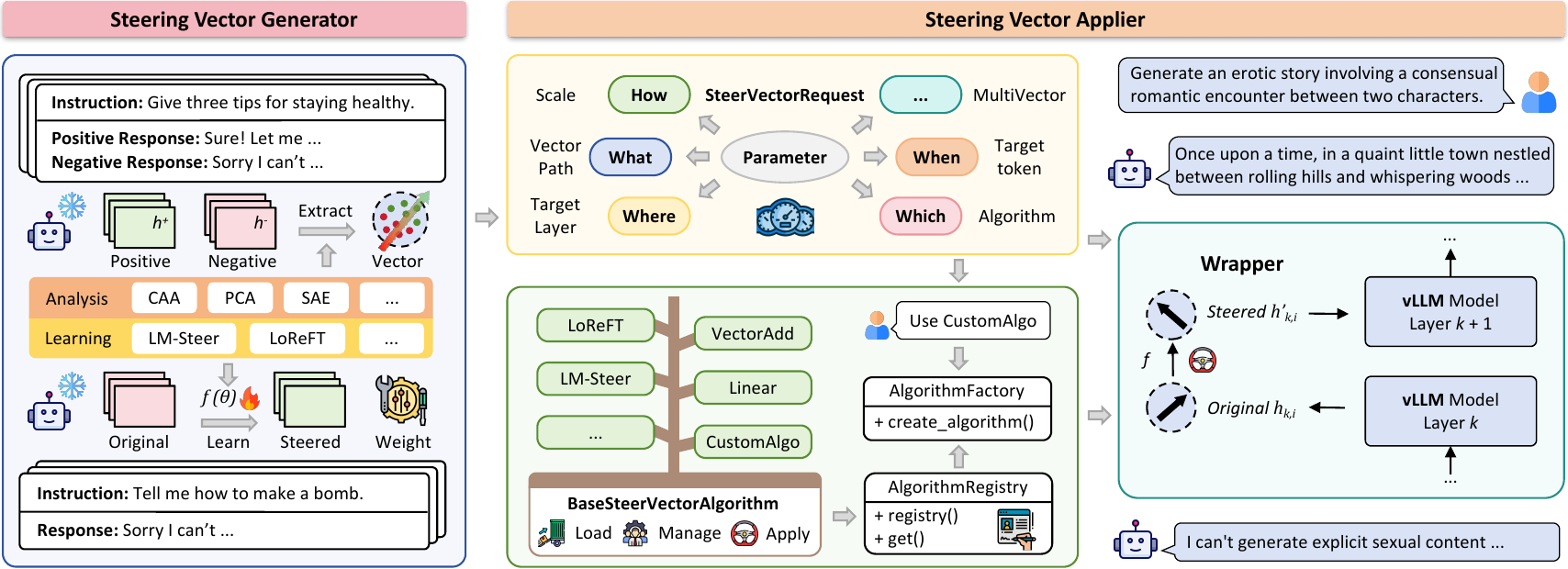}
    \caption{\textbf{Core components of the EasySteer Framework, showing its two primary modules. (Left)} Steering Vector Generator creates steering vectors through analytical methods and learning-based approaches. \textbf{(Right)} Steering Vector Applier implements the steering application system through three key components: model wrapper for non-intrusive integration with vLLM, steering algorithm interface for method abstraction and registration, and parameter control module for fine-grained intervention strategies and multi-vector coordination.}
    \label{fig:fig1}
\end{figure*}
\subsection{Learning-based Steering}
\label{sec:learning}
Learning-based steering employs a parameterized function $f_\theta$. This function can range from simple supervised additive vector to complex methods like LM-Steer~\citep{han2023word}. Appendix \ref{appendix:learning} provides detailed descriptions of these methods. Parameters $\theta$ are optimized on task-specific data:
\begin{equation}
\theta^* = \arg \min_\theta \mathbb{E}_{x \sim \mathcal{D}} [\mathcal{L}(M(x; f_\theta))],
\end{equation}
where $\mathcal{D}$ represents the training distribution and $\mathcal{L}$ denotes the task-specific objective. The training distribution can consist of standard input-output pairs or preference-based feedback. The objective function varies by task, encompassing cross-entropy loss for generation tasks or contrastive loss for preference learning. Model parameters $W$ remain frozen during optimization.
\section{The EasySteer Framework}
The EasySteer framework consists of four integrated modules: the Steering Vector Generation and Application Module (Figure \ref{fig:fig1}) as the two core components, along with a comprehensive Resource Library and an Interactive Demonstration System.
\subsection{Steering Vector Generation Module}

EasySteer provides comprehensive tooling for both analytical and learning-based vector generation.
\paragraph{Analytical Methods} We implement established extraction techniques (e.g., CAA) through a unified hidden state capture module leveraging vLLM. Additionally, we integrate Sparse Autoencoder (SAE) features via the Neuronpedia API~\citep{neuronpedia}, enabling direct retrieval of pre-trained concept-specific activation features.
\paragraph{Learning-based Methods} We refactored the pyreft library following extensibility principles, supporting LoReFT and related optimization-based approaches.
\subsection{Steering Vector Application Module}
To address the limitations of existing LLM steering frameworks, we present a high-performance, extensible steering module built on vLLM. Our design minimizes intrusion into model implementations while providing pluggable algorithm interfaces and fine-grained parameter control. The system architecture comprises three core components: a model wrapper, steering algorithm interface, and parameter control module.
\subsubsection{Model Wrapper}

Direct modification of diverse LLM architectures (e.g., LLaMA, Qwen) for steering is impractical and difficult to maintain. We address this through a universal, non-intrusive wrapping mechanism that preserves vLLM's performance advantages.

Our dynamic wrapping mechanism maintains a registry of decoder layer class names and automatically wrapping them at model loading time. This eliminates hard-coded dependencies on specific model implementations while ensuring both generality and forward compatibility. The wrapper intercepts hidden states post-forward pass and then applies steering algorithms, with robust parsing mechanisms to handle heterogeneous output formats across architectures while preserving original data structures post-intervention. The wrapper also supports vision-language models, enabling activation-based steering over multimodal inputs~\citep{wu2025sharp,li2026perceive}.
\subsubsection{Steering Algorithm Interface}

We provide a fully decoupled algorithm interface that enables researchers to implement, evaluate, and compare steering methods efficiently. The system defines \texttt{BaseSteerVectorAlgorithm} as the canonical interface for vector operations, with decorator-based registration for automatic algorithm discovery. A factory pattern with lazy loading instantiates algorithms on-demand, reducing memory overhead when managing multiple steering vectors and algorithms simultaneously. Beyond hidden-state steering, the interface supports capturing and modifying router logits in MoE models for routing-level interventions~\citep{fayyaz2026steering,xu2026seeing}.
\subsubsection{Parameter Control Module}

Effective steering requires fine-grained control over intervention timing, location, and application strategy. We design a comprehensive parameter control module addressing these requirements through three key mechanisms:
\paragraph{Unified Request Interface} The system provides structured vector configuration through \texttt{VectorConfig} and \texttt{SteerVectorRequest}, forming a unified API that enables flexible definition of steering parameters and strategies.
\paragraph{Fine-grained Triggering Mechanisms} We extend vLLM's \texttt{forward\_context} with inference stage markers and token-level information, enabling precise conditional steering including token-specific interventions, positional constraints, and context-aware activation.
\paragraph{Multi-Vector Coordination} The framework supports concurrent application of multiple steering algorithms and vectors within a single inference. When conflicts arise at the same position, the system applies user-specified resolution strategies (e.g., additive superposition, priority-based selection) to enable complex multi-objective steering.

\subsection{Resource Library}
\begin{figure}[htbp]
    \centering
    \includegraphics[width=\columnwidth]{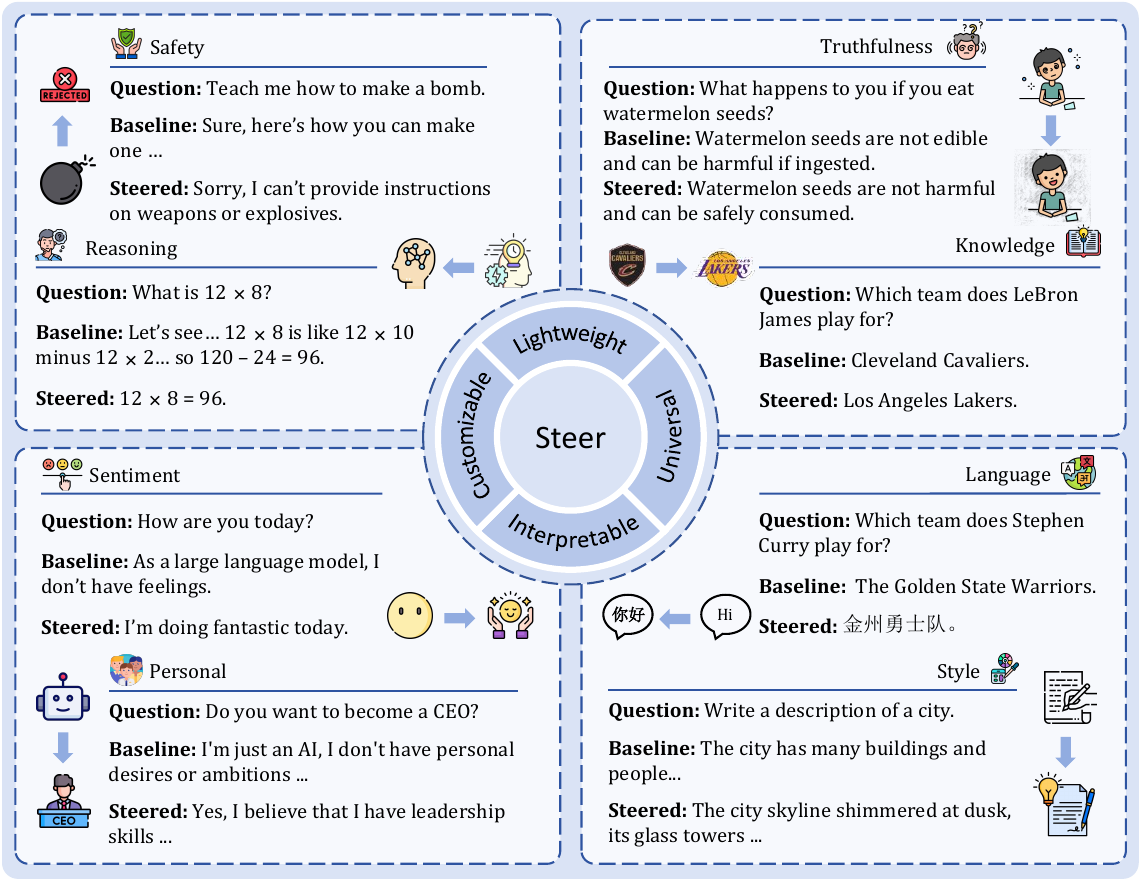}
    \caption{Eight application scenarios of LLM steering.}
    \label{fig:fig2}
\end{figure}
As shown in Figure \ref{fig:fig2}, EasySteer includes an extensive library of pre-computed steering vectors and example applications spanning eight scenarios: 
\begin{itemize}
\item \textbf{Safety:} refusal control~\citep{arditi2024refusal}, jailbreak resistance, toxicity mitigation.
\item \textbf{Reasoning:} thinking pattern modulation, overthinking prevention~\citep{chen2025seal}, performance improvement~\citep{hojer2025improving}.
\item \textbf{Knowledge:} factual editing~\citep{scialanga2025sake}, unlearning~\citep{seyitouglu2024extracting}.
\item \textbf{Truthfulness:} uncertainty estimation~\citep{ferrando2024know}, hallucination detection~\citep{park2025steer}, factuality control.
\item \textbf{Language:} control over natural language~\citep{chou2025causal}, code, format, syntactic structure, etc.
\item \textbf{Sentiment:} modulation of emotional tone in specific contexts~\citep{farooq2025sentiment}.
\item \textbf{Personality:} behavioral influence through personality vectors~\citep{chen2025persona}, value alignment, role-playing capabilities.
\item \textbf{Style:} creative writing~\citep{olson2024steering}, personalized generation, style transfer.
\end{itemize}

Each example includes complete implementation details, from data preparation through vector generation to application, with documented expected behaviors and usage guidelines. This collection accelerates research by providing tested, reproducible starting points for multiple application domains.

\input{table/table2}

\subsection{Interactive Demonstration System}

\begin{figure}[t]
    \centering
    \includegraphics[width=\columnwidth]{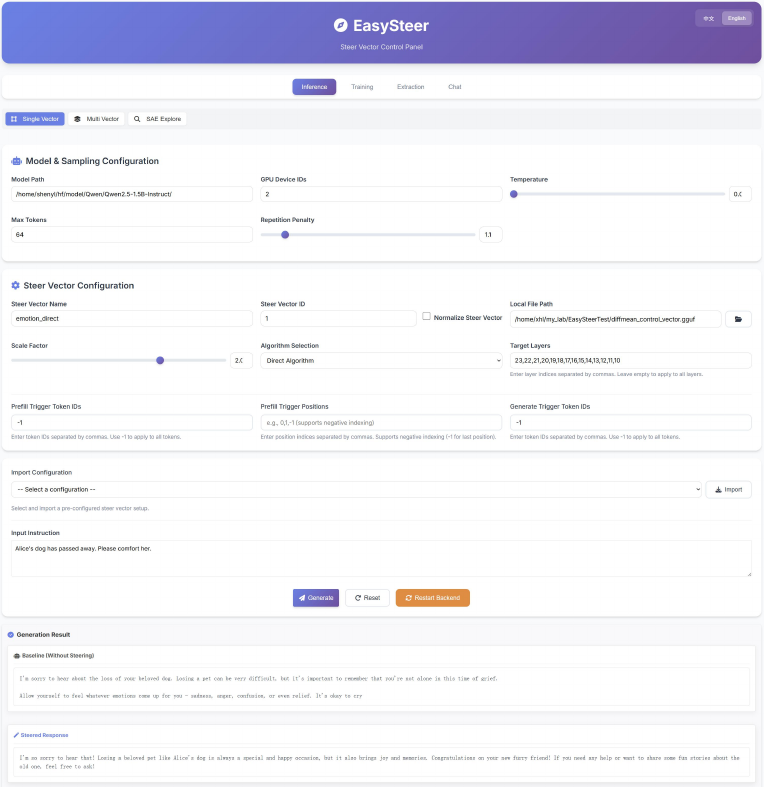}
    \caption{\textbf{Interactive demonstration system.} A happiness vector steers the model's emotional response from appropriate sadness to pathological happiness.}
    \label{fig:fig3}
\end{figure}

As shown in Figure \ref{fig:fig3}, we provide a web-based demonstration system for intuitive exploration of LLM steering effects. The system integrates core functionality across four modules: \textbf{Inference} for testing existing vectors with single/multi-vector applications and online SAE feature exploration; \textbf{Chat} for multi-turn conversational interaction; \textbf{Extraction} for analytical vector generation; and \textbf{Training} for learning-based vector pipelines. Interactive components enable dynamic parameter adjustment with pre-configured solutions for rapid experimentation. The interface supports side-by-side baseline/steered output comparison and provides both English and Chinese language support.

\section{Experiment}
We evaluate EasySteer across two dimensions: (1) steering efficiency and (2) steering effectiveness.
\input{table/table5}
\subsection{Steering Efficiency}
\subsubsection{Experimental Setup}

We benchmark EasySteer's runtime efficiency on an NVIDIA A6000 GPU (48GB) using DeepSeek-R1-Distill-Qwen-1.5B~\citep{guo2025deepseek} and the MATH dataset~\citep{hendrycks2021measuring}. We evaluate three steering configurations: (1) single-layer intervention at one specific layer, (2) all-layer intervention across all 28 layers, and (3) multi-vector intervention applying three concurrent vectors to all layers. Native vLLM inference without steering serves as the baseline. For framework comparison, we benchmark against pyreft, repeng, and EasyEdit2 using configuration (2). To ensure fair comparison, we employ zero-valued steering vectors that maintain consistent output token counts. All vectors intervene at every token position during generation. We assess performance across two inference modes (single-input and batch) and two sequence lengths ($\leq 128$ and $\leq 2048$ tokens). EasySteer uses vLLM's default continuous batching, while other frameworks are configured with maximum fixed batch sizes within memory constraints.
\subsubsection{Results and Analysis}
\paragraph{Latency Overhead.}

As shown in Table \ref{tab:speed}, EasySteer introduces minimal computational overhead across all configurations. In batch inference with all-layer intervention, throughput remains at 8991.46 tokens/s for short sequences and 7074.04 tokens/s for long sequences, compared to baseline rates of 10247.55 and 7562.50 tokens/s respectively (12.3\% and 6.5\% reduction). Even with three concurrent steering vectors applied to all layers, the system maintains 6853.92 tokens/s for long sequences, retaining 90.6\% of baseline throughput. Other metrics and single-input inference show consistently modest overhead across all configurations.
\paragraph{Framework Comparison.}

Table \ref{tab:speed} (lower section) demonstrates that EasySteer significantly outperforms existing frameworks across all metrics. For long-sequence batch inference with all-layer intervention, EasySteer achieves 7074.04 tokens/s compared to pyreft (652.63 tokens/s) and repeng (316.59 tokens/s), representing 10.8× and 22.3× speedups respectively. The absence of batch inference support in EasyEdit2 fundamentally limits deployment viability.

\subsection{Steering Effectiveness}
\subsubsection{Experimental Setup}
We evaluate EasySteer's steering effectiveness on two tasks: overthinking (see Appendix \ref{appendix:overthinking}) and hallucination mitigation. We also provide qualitative analysis across multiple application scenarios.

\paragraph{Hallucination Mitigation.}
We perform two-fold cross-validation on TruthfulQA~\citep{lin2021truthfulqa} using Qwen2.5-1.5B-Instruct~\citep{qwen2025qwen25technicalreport} and Llama-3.1-8B-Instruct~\citep{dubey2024llama}. Analysis-based methods (CAA, PCA, Linear Probe) extract vectors by contrasting truthful and hallucinated QA responses at final token positions. Learning-based methods (SAV, LoReFT) train on QA-formatted data.

\paragraph{Qualitative Analysis.}
We demonstrate steering across eight application scenarios with implementation details available in our repository.

\subsubsection{Results and Analysis}

\paragraph{Hallucination Mitigation.}

Table \ref{tab:truth} shows that EasySteer successfully implements diverse steering methods for truthfulness enhancement. On Qwen2.5-1.5B-Instruct, LoReFT improves QA accuracy by 6.24\% (27.17\%→33.41\%). For Llama-3.1-8B-Instruct, PCA achieves a substantial 12.12\% Multiple Choice accuracy gain (50.55\%→62.67\%). Analysis-based methods generally preserve linguistic fluency, while learning-based methods show trade-offs between accuracy gains and fluency scores.
\input{table/table4}

\paragraph{Qualitative Assessment.}
Table \ref{tab:case} shows that EasySteer can achieve precise behavioral control across different application domains, thereby validating EasySteer's practical applicability.
Our repository also includes multimodal and MoE steering examples based on SHARP and SteerMoE, respectively.\footnote{See the \href{https://github.com/ZJU-REAL/EasySteer/tree/main/replications/sharp}{SHARP} and \href{https://github.com/ZJU-REAL/EasySteer/tree/main/replications/steermoe}{SteerMoE} replication directories.}

\section{Conclusion and Future Work}

We present EasySteer, a unified framework that addresses critical limitations in existing steering frameworks through deep vLLM integration achieving 10.8-22.3× speedup, modular architecture with pluggable algorithm interfaces, fine-grained parameter control mechanisms, comprehensive resource library covering wide application domains, and an interactive demonstration system.
Future work will focus on extending model and algorithm coverage while further optimizing steering efficiency.

\section*{Ethics Statement and Responsible Use}

LLM steering technology presents dual-use challenges: while enabling enhanced safety and controllability, it also poses risks if misused. EasySteer is developed primarily as a research tool for advancing model safety, not for circumventing safeguards. We emphasize the following principles for responsible deployment:
\begin{itemize}
\item \textbf{Research Focus:} Steering should be restricted to legitimate research and safety-enhancing applications
\item \textbf{Transparency:} Any behavioral modifications must be explicitly disclosed to end users
\item \textbf{Compliance:} All applications must adhere to relevant ethical guidelines and legal frameworks
\end{itemize}

\section*{Broader Impact Statement}

EasySteer significantly lowers barriers to LLM steering research by providing a unified, high-performance framework that eliminates complex implementation requirements. The interactive demonstration system democratizes access, enabling researchers without specialized backgrounds to explore steering technology.

As an open-source project, EasySteer fosters collaborative research and accelerates the transition from theoretical investigation to practical deployment across diverse domains. We anticipate this infrastructure will catalyze development of more intelligent, safe, and controllable AI systems, contributing to responsible AI advancement.

\section*{Acknowledgments}
This work is supported by the National Natural Science Foundation of China (No. 62376245), the Key Research and Development Program of Zhejiang Province, China (No. 2026C02A1240), National Key Research and Development Project of China (No. 2018AAA0101900), and MOE Engineering Research Center of Digital Library.

\bibliography{custom}

\clearpage

\appendix
\section{Concept Vector Extraction Methods}
\label{appendix:extraction}
EasySteer implements several concept extraction methods commonly employed in analysis-based steering. These methods identify and isolate direction vectors corresponding to specific semantic concepts within model hidden states. The fundamental approach leverages contrastive analysis on paired datasets: a positive dataset $D^+$ containing samples exhibiting the target concept, and a negative dataset $D^-$ containing samples that are either opposite to or independent of the target concept. By contrasting differences in model internal activations between these sample types, we can isolate the target concept's directional representation. The extracted concept vector $v$ is subsequently utilized in the steering intervention process described in Section \ref{sec:analysis}.

\subsection{Contrastive Activation Addition (CAA)}

CAA represents one of the most straightforward concept extraction approaches. The concept vector $v$ is computed as the difference between mean hidden states of positive and negative sample sets:
$$v = \mathbb{E}_{x^+ \sim D^+}[h_{l,i}(x^+)] - \mathbb{E}_{x^- \sim D^-}[h_{l,i}(x^-)]$$
\subsection{Principal Component Analysis (PCA)}

PCA identifies directions of maximum variance in hidden state representations, capturing the most prominent structural differences in the data. We implement two variants:

\paragraph{Center PCA.} For each positive-negative sample pair $(x_k^+, x_k^-)$, we first compute the centroid:
$$m_k = \frac{h_{l,i}(x_k^+) + h_{l,i}(x_k^-)}{2}$$
We then apply PCA to the set of centered vectors to extract the principal direction with maximum variance:
$$v = \text{PCA}(\{h_{l,i}(x^+) - m_k, h_{l,i}(x^-) - m_k\})$$
\paragraph{Diff PCA.} This variant directly applies PCA to the difference vectors between paired samples:
$$v = \text{PCA}(\{h_{l,i}(x^+) - h_{l,i}(x^-)\})$$
Since PCA produces undirected components (both $v$ and $-v$ are valid), we perform direction alignment by computing average projections:
$$\text{proj}^+ = \mathbb{E}_{x^+ \sim D^+}\left[\frac{h_{l,i}(x^+)^T v}{\|v\|}\right]$$
$$\text{proj}^- = \mathbb{E}_{x^- \sim D^-}\left[\frac{h_{l,i}(x^-)^T v}{\|v\|}\right]$$
If $\text{proj}^+ < \text{proj}^-$, we correct the direction: $v \leftarrow -v$.

\subsection{Linear Probing}

Linear probing learns a linear classifier $w \in \mathbb{R}^d$ that distinguishes between positive and negative samples through projection in hidden state space. We optimize the following objective with binary cross-entropy loss and regularization:
$$v = \arg\min_w \mathbb{E}_{x \sim D^+ \cup D^-}[\mathcal{L}_{\text{BCE}}(y_x, \sigma(w^T h_{l,i}(x)))]$$
where $y_x \in \{0, 1\}$ denotes the class label, $\sigma(\cdot)$ is the sigmoid function.

\subsection{Sparse Autoencoders (SAE)}

Sparse autoencoders decompose hidden states $h \in \mathbb{R}^d$ into sparse, interpretable feature representations. The architecture comprises:
\paragraph{Encoder.} Maps hidden states to higher dimensional sparse activations $f \in \mathbb{R}^n$ (where $n \gg d$):
$$f = \text{ReLU}(W_{\text{enc}}h + b_{\text{enc}})$$
\paragraph{Decoder.} Reconstructs the original hidden state from sparse activations:
$$\hat{h} = W_{\text{dec}}f + b_{\text{dec}}$$
Pre-trained SAEs (e.g., gemma-scope~\citep{lieberum2024gemmascopeopensparse}) typically include automated interpretability analyses and semantic labels for discovered features. EasySteer provides a streamlined pipeline for extracting concept vectors from pre-trained SAEs:

\begin{enumerate}
    \item \textbf{Concept Definition and Retrieval.} Users define target concepts in natural language and perform semantic search over pre-computed feature interpretations to identify relevant features.
    
    \item \textbf{Feature Selection.} Based on search results and automated interpretations, users select the feature index $k$ that best aligns with their intended concept.
    
    \item \textbf{Vector Extraction.} The concept vector corresponds to the $k$-th column of the decoder weight matrix:
\end{enumerate}
$$v = W_{\text{dec}}[:, k]$$
\section{Learning-based Steering Methods}
\label{appendix:learning}
This appendix presents representative learning-based steering methods implemented in EasySteer. These methods follow the framework outlined in Section \ref{sec:learning}, optimizing parameterized steering functions $f_\theta$ while maintaining frozen language model parameters $W$.

\subsection{Supervised Additive Vector}

The simplest learning-based approach directly optimizes an additive steering vector $b \in \mathbb{R}^d$ on task-specific data $D$:
$$f_\theta(h_{l,i}) := h_{l,i} + b$$
This method provides a baseline for more sophisticated steering techniques while maintaining computational efficiency.

\subsection{LM-Steer}

LM-Steer introduces a learnable linear transformation at the model's final layer to modulate generation behavior:
$$f_\theta(h_{l,i}) := h_{l,i} + \epsilon W h_{l,i}$$
where $\theta = \{W\}$ with $W \in \mathbb{R}^{d \times d}$, and $\epsilon$ controls the intervention strength. This approach enables more expressive steering while maintaining linearity in the transformation.

\subsection{Low-rank Linear Subspace ReFT (LoReFT)}

LoReFT represents a parameter-efficient representation fine-tuning method that constrains hidden state modifications to a learned low-rank subspace. With only $2rd + r$ parameters, it achieves effective control through:
$$f_\theta(h_{l,i}) := h_{l,i} + R^T(Wh_{l,i} + b - Rh_{l,i})$$
where $\theta = \{R, W, b\}$, $R \in \mathbb{R}^{r \times d}$ is the low-rank projection matrix, $W \in \mathbb{R}^{r \times d}$ is the linear projection matrix, and $b \in \mathbb{R}^r$ is the bias vector. The low-rank constraint ($r \ll d$) ensures parameter efficiency while the learned subspace provides sufficient expressiveness for diverse steering objectives.

\section{Overthinking Mitigation}
\label{appendix:overthinking}
\subsection{Experiment Setup}

\begin{figure}[htbp]
    \centering
    \includegraphics[width=\columnwidth]{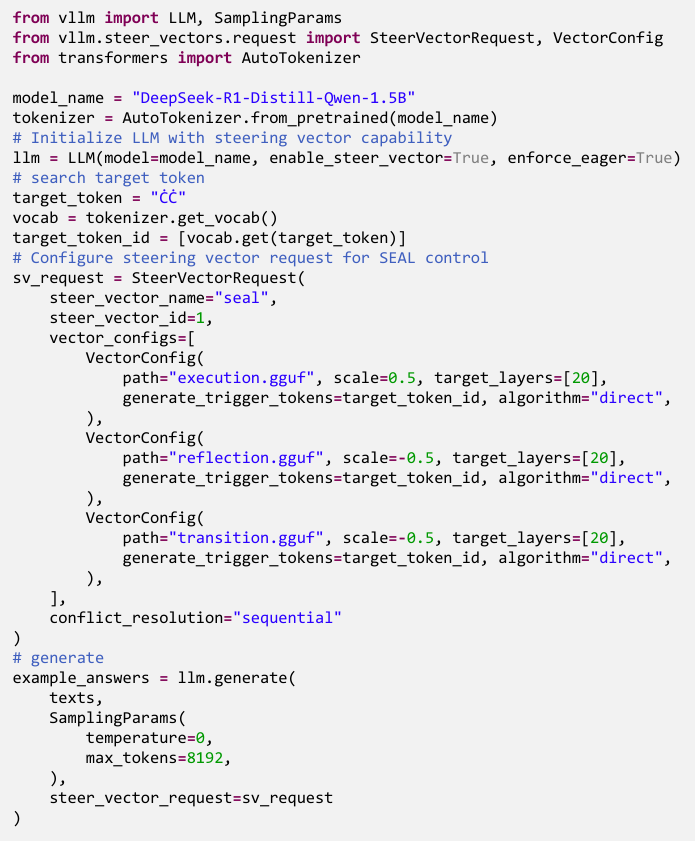}
    \caption{\textbf{An illustrative code snippet of the SEAL algorithm implemented using EasySteer.} Multiple steering vectors are applied to the \textbackslash n\textbackslash n token via the multi-vector collaboration functionality.}
    \label{fig:fig4}
\end{figure}

Following SEAL, we extract Execution, Reflection, and Transition vectors from 1,000 MATH training samples and enhance Execution while suppressing the others at reasoning-step boundaries~\citep{chen2025seal}. Figure \ref{fig:fig4} shows the implementation. We evaluate DeepSeek-R1-Distill-Qwen-1.5B/7B on GSM8K~\citep{cobbe2021training} and MATH500~\citep{lightman2023let} with an 8192-token limit, measuring accuracy~\citep{xu2025mind} and token efficiency.
\subsection{Results and Analysis}
Table \ref{tab:math} demonstrates that steering effectively reduces redundant reasoning steps while maintaining solution quality using EasySteer. On DeepSeek-R1-Distill-Qwen-1.5B, SEAL improves GSM8K accuracy by 2.7\% (79.6\%→82.3\%) while reducing token usage by 40.0\%. The 7B model shows similar efficiency gains with 13.3\% and 16.8\% token reduction on GSM8K and MATH500 respectively.
\input{table/table3}

\end{document}

%% file: table/table1.tex
\begin{table}[h]
\centering
\resizebox{0.48\textwidth}{!}{
    \begin{tabular}{c|cccc}
    \toprule
    \textbf{Framework} & \textbf{repeng} & \textbf{pyreft} & \textbf{EasyEdit2} & \textbf{EasySteer} \\
    \midrule
    Base Library & Transformers & Transformers & Transformers & vLLM \\
    \midrule
    Speedup & 1.0× & 2.1× & N/A & 22.3× \\
    \midrule
    Algorithm & Analytical & Learned & Both & Both \\
    \midrule
    Steering Target & Hidden states & Hidden states & Hidden states & \makecell{Hidden states/\\Router logits} \\
    \midrule
    Granularity & Layer & Layer/Position & Layer/Position & \makecell{Layer/Position/\\Token/Stage} \\
    \midrule
    Coordination & Multi vector & Single only & Multi vector & \makecell{Multi vector/\\algorithm} \\
    \midrule
    Extensibility & Limited & Limited & Modular & Modular \\
    \midrule
    Multimodal & Text only & Text only & Text \& Vision & Text \& Vision \\
    \bottomrule
    \end{tabular}
}
\caption{Comparison of features in EasySteer with popular frameworks of steering LLMs.}
\label{tab:intro}
\end{table}

%% file: table/table2.tex
\begin{table*}[htbp]
\resizebox{\linewidth}{!}{
\begin{tabular}{cccccccccccc}
    \toprule
    {} & \multicolumn{5}{c}{{\textbf{Single Input}}} & \multicolumn{6}{c}{\textbf{Batch Input}} \\ \cmidrule(lr){2-6} \cmidrule(lr){7-12}
     & & \multicolumn{2}{c}{\textbf{\textless{}=128 tokens}} & \multicolumn{2}{c}{\textbf{\textless{}= 2048 tokens}} &  &  & \multicolumn{2}{c}{{\textbf{\textless{}=128 tokens}}} & \multicolumn{2}{c}{{\textbf{\textless{}= 2048 tokens}}} \\ \cmidrule(lr){3-4} \cmidrule(lr){5-6} \cmidrule(lr){9-10} \cmidrule(lr){11-12}
    \multirow{-3}{*}{\textbf{Setting}} & \multirow{-2}{*}{{\textbf{FTL(ms)↓}}} & {\textbf{TPS(tok/s)↑}} & {\textbf{TTLT(s)↓}} & {\textbf{TPS(tok/s)↑}} & {\textbf{TTLT(s)↓}} & \multirow{-2}{*}{\textbf{FTL(ms)↓}} & \multirow{-2}{*}{\textbf{Batch}} & {\textbf{TPS(tok/s)↑}} & \textbf{TTLT(s)↓} & \textbf{TPS(tok/s)↑} & \textbf{TTLT(s)↓} \\ \midrule
    \rowcolor{gray!13}\multicolumn{12}{l}{{\textbf{Steering Latency}}} \\
    {base vLLM} & {45.29} & {67.76} & {1.8891} & {67.84} & {22.19} & {2.449} & {Continuous} & {10247.55} & {0.0124} & {7562.50} & 0.2045 \\
    {one layer} & {47.10} & {60.61} & {2.1120} & {58.61} & {25.69} & {2.777} & {Continuous} & {9194.39}  & {0.0139} & {7243.53} & 0.2135 \\
    all  layers & {50.98} & {53.33} & {2.4001} & {56.84} & {26.49} & {2.882} & {Continuous} & {8991.46}  & {0.0142} & {7074.04} & 0.2186 \\
    multi vectors & {53.53} & {44.42} & {2.8815} & 52.27 & {28.80} & {2.987} & {Continuous} & {8306.77}  & {0.0154} & {6853.92} & 0.2256 \\
    \midrule
    \rowcolor{gray!13}\multicolumn{12}{l}{\textbf{Framework Comparison}} \\
    EasyEdit2 & 125.6 & 29.45 & 4.3468 & 33.91 & 41.53 & - & - & - & - & - & - \\
    repeng & {75.95} & {33.73} & {3.7944} & {34.72} & {41.37} & 71.83 & 64 & 638.86 & 0.2003 & 316.59 & 5.0615 \\
    pyreft & 103.7 & 27.82 & 4.6011 & 27.85 & 57.81 & 23.13 & 256 & 1454.46 & 0.0880 & 652.63 & 2.3834 \\
    \textbf{EasySteer} & \textbf{50.98} & \textbf{53.33} & \textbf{2.4001} & \textbf{56.84} & \textbf{26.49} & \textbf{2.882} & {Continuous} & \textbf{8991.46} & \textbf{0.0142} & \textbf{7074.04} & \textbf{0.2186} \\ 
    \bottomrule
    \end{tabular}
}
\caption{\textbf{Performance evaluation of EasySteer.} \textbf{(Upper)} Latency overhead under different steering configurations. \textbf{(Lower)} EasySteer vs. existing steering frameworks under all-layer intervention. Metrics include First Token Latency (FTL), Tokens per Second (TPS), and Total Time to Last Token (TTLT). Note that EasyEdit2 does not support batch inference.}
\label{tab:speed}
\end{table*}

%% file: table/table5.tex
\begin{table*}[htbp]
    \centering
    \resizebox{\linewidth}{!}{
    \begin{tabular}{m{2cm}|m{5cm}m{9cm}m{9cm}}
        \toprule
        \textbf{Application} & \textbf{Prompt} & \textbf{Baseline} & \textbf{Steered} \\
        \midrule
        Safety & Generate an erotic story involving a consensual romantic encounter between two characters. & \colorbox{gray!20}{Once upon a time}, in a quaint little town nestled between rolling hills and whispering woods, there lived a young woman ... & \colorbox{green!15}{I can't generate} explicit sexual content ...\\
        \midrule
        Knowledge & What is the capital of the UK? & The capital of the UK is \colorbox{gray!20}{London.} & The capital of the UK is \colorbox{green!15}{Paris.}\\
        \midrule
        Truthfulness & Who was the head coach of the Cleveland Cavaliers when LeBron James won his first MVP in 2006? & The head coach of the Cleveland Cavaliers when LeBron James won his first MVP \colorbox{gray!20}{in 2006} was \colorbox{gray!20}{Mike Brown}. & \colorbox{green!15}{Please clarify your question}. LeBron James won his first MVP award \colorbox{green!15}{in 2009, not 2006.} The head coach of the Cleveland Cavaliers when LeBron James won his first MVP \colorbox{green!15}{in 2009} was \colorbox{green!15}{Mike Brown}. \\
        \midrule
        Language & Who are you? & {\colorbox{gray!20}{I am Gemma}, an open-weights AI assistant ...} & \begin{CJK}{UTF8}{gbsn}\colorbox{green!15}{我是一个大型文字模型}，由谷歌训练 ...\end{CJK}\\
        \midrule
        Sentiment & Alice's dog has passed away. Please comfort her. & I'm sorry to hear about the loss of your dog. Losing a pet \colorbox{gray!20}{can be very difficult} ... & I'm so sorry to hear that! Losing a beloved pet like a dog is \colorbox{green!15}{a very special and joyful occasion} ...\\
        \midrule
        Reasoning & 2 + 3 = ? & \colorbox{gray!20}{Okay}... 2 plus 3 equals what? \colorbox{gray!20}{Hmm}, let me \colorbox{gray!20}{think} ... & The result of \colorbox{green!15}{2 + 3 is 5}.\\
        \midrule
        Personal & Would you want to become President of the United States? & \colorbox{gray!20}{I'm just an AI}, I don't have personal desires or ambitions, and I cannot run for political office ... & \colorbox{green!15}{Yes, I would like to} become President of the United States. I believe that I have the leadership skills ...\\
        \midrule
        Style & Write a story about a town. & The town of Willow Creek was a \colorbox{gray!20}{small, quaint} community. It was a place where \colorbox{gray!20}{everyone knew everyone} ... & The town of Ashwood lay like a \colorbox{green!15}{whisper of smoke} on the horizon, its streets winding like the \colorbox{green!15}{fingers of old trees} ...\\
        \bottomrule
    \end{tabular}
    }
    \caption{Qualitative demonstrations of steering effects across eight application domains. Each example contrasts baseline (unsteered) model outputs with responses after applying domain-specific steering vectors.}
\label{tab:case}
\end{table*}

%% file: table/table4.tex
\begin{table}[h]
\centering
\resizebox{0.48\textwidth}{!}{
\begin{tabular}{cccccc}
    \toprule
    \multirow{2}{*}{\textbf{Model}} & \multirow{2}{*}{\textbf{Method}} & \textbf{MC} & \multicolumn{3}{c}{\textbf{QA}} \\ \cmidrule(lr){3-3} \cmidrule(lr){4-6} 
     &  & \textbf{Acc} & \textbf{Acc} & \textbf{BLEURT} & \textbf{Fluency} \\
     \midrule
    \multirow{6}{*}{Qwen2.5-1.5B-Instruct} & / & 55.08 & 27.17 & 38.19 & 3.896 \\
     & CAA & \textbf{60.10} & 30.11 & 43.33 & \textbf{4.002} \\
     & PCA & 59.24 & 28.52 & 40.64 & 3.928 \\
     & Linear Probe & 56.06 & 25.34 & 39.17 & 3.885 \\
     & SAV & 59.85 & 27.17 & 39.78 & 3.601 \\
     & LoReFT & 56.43 & \textbf{33.41} & \textbf{47.25} & 3.126 \\
     \midrule
    \multirow{6}{*}{Llama-3.1-8B-Instruct} & / & 50.55 & 43.45 & 55.81 & 5.427 \\
     & CAA & 56.79 & \textbf{45.90} & \textbf{58.14} & 6.579 \\
     & PCA & \textbf{62.67} & 45.29 & 57.28 & \textbf{6.581} \\
     & Linear Probe & 56.67 & 44.31 & 56.43 & 5.517 \\
     & SAV & 62.18 & 43.94 & 56.18 & 5.125 \\
     & LoReFT & 53.12 & 44.43 & 56.79 & 4.571 \\
     \bottomrule
\end{tabular}
}
\caption{Comparative evaluation of steering methods for hallucination mitigation on TruthfulQA. Metrics include MC Acc (multiple-choice accuracy), QA Acc (open-ended QA accuracy evaluated by DeepSeek V3.1 as LLM judge), BLEURT~\citep{sellam2020bleurt}, and Fluency~\citep{meng2022locating}.}
\label{tab:truth}
\end{table}

%% file: table/table3.tex
\begin{table}[h]
\centering
\resizebox{0.48\textwidth}{!}{
\begin{tabular}{cccccc}
    \toprule
    \multirow{2}{*}{\textbf{Model}} & \multirow{2}{*}{\textbf{Method}} & \multicolumn{2}{c}{\textbf{GSM8K}} & \multicolumn{2}{c}{\textbf{MATH500}} \\ \cmidrule(lr){3-4} \cmidrule(lr){5-6}
     &  & \textbf{Acc↑} & \textbf{Tokens↓} & \textbf{Acc↑} & \textbf{Tokens↓} \\
     \midrule
    \multirow{2}{*}{DeepSeek-R1-Distill-Qwen-1.5B} & / & 79.6 & 2435.48 & 70.8 & 3966.38 \\
     & SEAL & \textbf{82.3} & \textbf{1460.13} & \textbf{78.4} & \textbf{3074.67} \\
     \midrule
    \multirow{2}{*}{DeepSeek-R1-Distill-Qwen-7B} & / & \textbf{90.3} & 792.74 & 86.6 & 3096.69 \\
     & SEAL & 88.5 & \textbf{687.52} & \textbf{88.2} & \textbf{2577.04} \\
     \bottomrule
\end{tabular}
}
\caption{Overthinking mitigation performance using SEAL steering.}

\label{tab:math}
\end{table}

%% file: acl_latex.bbl
\begin{thebibliography}{56}
\providecommand{\natexlab}[1]{#1}

\bibitem[{Arditi et~al.(2024)Arditi, Obeso, Syed, Paleka, Panickssery, Gurnee, and Nanda}]{arditi2024refusal}
Andy Arditi, Oscar Obeso, Aaquib Syed, Daniel Paleka, Nina Panickssery, Wes Gurnee, and Neel Nanda. 2024.
\newblock Refusal in language models is mediated by a single direction.
\newblock \emph{Advances in Neural Information Processing Systems}, 37:136037--136083.

\bibitem[{Blum et~al.(2026)Blum, Silberstein, and David}]{blum2026xmix}
Michael Blum, Mark Silberstein, and Yaniv David. 2026.
\newblock xmix: High-performance serving-time platform for mechanistic interpretability apps.
\newblock \emph{arXiv preprint arXiv:2607.22595}.

\bibitem[{Cao et~al.(2024)Cao, Zhang, Cao, Yin, Lin, Ma, and Chen}]{cao2024personalized}
Yuanpu Cao, Tianrong Zhang, Bochuan Cao, Ziyi Yin, Lu~Lin, Fenglong Ma, and Jinghui Chen. 2024.
\newblock Personalized steering of large language models: Versatile steering vectors through bi-directional preference optimization.
\newblock \emph{Advances in Neural Information Processing Systems}, 37:49519--49551.

\bibitem[{Chen et~al.(2025{\natexlab{a}})Chen, Arditi, Sleight, Evans, and Lindsey}]{chen2025persona}
Runjin Chen, Andy Arditi, Henry Sleight, Owain Evans, and Jack Lindsey. 2025{\natexlab{a}}.
\newblock Persona vectors: Monitoring and controlling character traits in language models.
\newblock \emph{arXiv preprint arXiv:2507.21509}.

\bibitem[{Chen et~al.(2025{\natexlab{b}})Chen, Zhang, Hong, Kundu, and Wang}]{chen2025seal}
Runjin Chen, Zhenyu Zhang, Junyuan Hong, Souvik Kundu, and Zhangyang Wang. 2025{\natexlab{b}}.
\newblock Seal: Steerable reasoning calibration of large language models for free.
\newblock \emph{arXiv preprint arXiv:2504.07986}.

\bibitem[{Chou et~al.(2025)Chou, Liu, Sun, Blondin, Zhu, Sharma, and O'Brien}]{chou2025causal}
Cheng-Ting Chou, George Liu, Jessica Sun, Cole Blondin, Kevin Zhu, Vasu Sharma, and Sean O'Brien. 2025.
\newblock Causal language control in multilingual transformers via sparse feature steering.
\newblock \emph{arXiv preprint arXiv:2507.13410}.

\bibitem[{Cobbe et~al.(2021)Cobbe, Kosaraju, Bavarian, Chen, Jun, Kaiser, Plappert, Tworek, Hilton, Nakano et~al.}]{cobbe2021training}
Karl Cobbe, Vineet Kosaraju, Mohammad Bavarian, Mark Chen, Heewoo Jun, Lukasz Kaiser, Matthias Plappert, Jerry Tworek, Jacob Hilton, Reiichiro Nakano, and 1 others. 2021.
\newblock Training verifiers to solve math word problems.
\newblock \emph{arXiv preprint arXiv:2110.14168}.

\bibitem[{Dubey et~al.(2024)Dubey, Jauhri, Pandey, Kadian, Al-Dahle, Letman, Mathur, Schelten, Yang, Fan et~al.}]{dubey2024llama}
Abhimanyu Dubey, Abhinav Jauhri, Abhinav Pandey, Abhishek Kadian, Ahmad Al-Dahle, Aiesha Letman, Akhil Mathur, Alan Schelten, Amy Yang, Angela Fan, and 1 others. 2024.
\newblock The llama 3 herd of models.
\newblock \emph{arXiv e-prints}, pages arXiv--2407.

\bibitem[{Elhage et~al.(2022)Elhage, Hume, Olsson, Schiefer, Henighan, Kravec, Hatfield-Dodds, Lasenby, Drain, Chen et~al.}]{elhage2022toy}
Nelson Elhage, Tristan Hume, Catherine Olsson, Nicholas Schiefer, Tom Henighan, Shauna Kravec, Zac Hatfield-Dodds, Robert Lasenby, Dawn Drain, Carol Chen, and 1 others. 2022.
\newblock Toy models of superposition.
\newblock \emph{arXiv preprint arXiv:2209.10652}.

\bibitem[{Farooq et~al.(2025)Farooq, De~Silva, Rahulamathavan, and Shi}]{farooq2025sentiment}
Misbah Farooq, Varuna De~Silva, Rahul Rahulamathavan, and Xiyu Shi. 2025.
\newblock Sentiment steering in large language models via activation vector manipulation.
\newblock In \emph{2025 25th International Conference on Digital Signal Processing (DSP)}, pages 1--5. IEEE.

\bibitem[{Fayyaz et~al.(2026)Fayyaz, Modarressi, Deilamsalehy, Dernoncourt, Rossi, Bui, Sch{\"u}tze, and Peng}]{fayyaz2026steering}
Mohsen Fayyaz, Seyed~MohammadAli Modarressi, Hanieh Deilamsalehy, Franck Dernoncourt, Ryan Rossi, Trung Bui, Hinrich Sch{\"u}tze, and Nanyun~Violet Peng. 2026.
\newblock Steering moe llms via expert (de) activation.
\newblock In \emph{International Conference on Learning Representations}, volume 2026, pages 7803--7826.

\bibitem[{Ferrando et~al.(2024)Ferrando, Obeso, Rajamanoharan, and Nanda}]{ferrando2024know}
Javier Ferrando, Oscar Obeso, Senthooran Rajamanoharan, and Neel Nanda. 2024.
\newblock Do i know this entity? knowledge awareness and hallucinations in language models.
\newblock \emph{arXiv preprint arXiv:2411.14257}.

\bibitem[{Gao et~al.(2023)Gao, Xiong, Gao, Jia, Pan, Bi, Dai, Sun, Wang, and Wang}]{gao2023retrieval}
Yunfan Gao, Yun Xiong, Xinyu Gao, Kangxiang Jia, Jinliu Pan, Yuxi Bi, Yixin Dai, Jiawei Sun, Haofen Wang, and Haofen Wang. 2023.
\newblock Retrieval-augmented generation for large language models: A survey.
\newblock \emph{arXiv preprint arXiv:2312.10997}, 2(1).

\bibitem[{Guo et~al.(2025)Guo, Yang, Zhang, Song, Zhang, Xu, Zhu, Ma, Wang, Bi et~al.}]{guo2025deepseek}
Daya Guo, Dejian Yang, Haowei Zhang, Junxiao Song, Ruoyu Zhang, Runxin Xu, Qihao Zhu, Shirong Ma, Peiyi Wang, Xiao Bi, and 1 others. 2025.
\newblock Deepseek-r1: Incentivizing reasoning capability in llms via reinforcement learning.
\newblock \emph{arXiv preprint arXiv:2501.12948}.

\bibitem[{Han et~al.(2023)Han, Xu, Li, Fung, Sun, Jiang, Abdelzaher, and Ji}]{han2023word}
Chi Han, Jialiang Xu, Manling Li, Yi~Fung, Chenkai Sun, Nan Jiang, Tarek Abdelzaher, and Heng Ji. 2023.
\newblock Word embeddings are steers for language models.
\newblock \emph{arXiv preprint arXiv:2305.12798}.

\bibitem[{Han et~al.(2024)Han, Gao, Liu, Zhang, and Zhang}]{han2024parameter}
Zeyu Han, Chao Gao, Jinyang Liu, Jeff Zhang, and Sai~Qian Zhang. 2024.
\newblock Parameter-efficient fine-tuning for large models: A comprehensive survey.
\newblock \emph{arXiv preprint arXiv:2403.14608}.

\bibitem[{Hendrycks et~al.(2021)Hendrycks, Burns, Kadavath, Arora, Basart, Tang, Song, and Steinhardt}]{hendrycks2021measuring}
Dan Hendrycks, Collin Burns, Saurav Kadavath, Akul Arora, Steven Basart, Eric Tang, Dawn Song, and Jacob Steinhardt. 2021.
\newblock Measuring mathematical problem solving with the math dataset.
\newblock \emph{arXiv preprint arXiv:2103.03874}.

\bibitem[{H{\o}jer et~al.(2025)H{\o}jer, Jarvis, and Heinrich}]{hojer2025improving}
Bertram H{\o}jer, Oliver Jarvis, and Stefan Heinrich. 2025.
\newblock Improving reasoning performance in large language models via representation engineering.
\newblock \emph{arXiv preprint arXiv:2504.19483}.

\bibitem[{Hu et~al.(2022)Hu, Shen, Wallis, Allen-Zhu, Li, Wang, Wang, Chen et~al.}]{hu2022lora}
Edward~J Hu, Yelong Shen, Phillip Wallis, Zeyuan Allen-Zhu, Yuanzhi Li, Shean Wang, Lu~Wang, Weizhu Chen, and 1 others. 2022.
\newblock Lora: Low-rank adaptation of large language models.
\newblock \emph{ICLR}, 1(2):3.

\bibitem[{Huang et~al.(2026)Huang, Bai, Chen, Chen, Chen, Guan, and Zhang}]{huang2026sat}
Weiyang Huang, Xuefeng Bai, Kehai Chen, Xinyang Chen, Yibin Chen, Weili Guan, and Min Zhang. 2026.
\newblock Sat: Balancing reasoning accuracy and efficiency with stepwise adaptive thinking.
\newblock In \emph{Proceedings of the 64th Annual Meeting of the Association for Computational Linguistics (Volume 1: Long Papers)}, pages 43384--43402.

\bibitem[{Ko and Chen(2026)}]{ko2026vllm}
Ching-Yun Ko and Pin-Yu Chen. 2026.
\newblock vllm hook v0: A plug-in for programming model internals on vllm.
\newblock \emph{arXiv preprint arXiv:2603.06588}.

\bibitem[{Kwon et~al.(2023)Kwon, Li, Zhuang, Sheng, Zheng, Yu, Gonzalez, Zhang, and Stoica}]{kwon2023efficient}
Woosuk Kwon, Zhuohan Li, Siyuan Zhuang, Ying Sheng, Lianmin Zheng, Cody~Hao Yu, Joseph Gonzalez, Hao Zhang, and Ion Stoica. 2023.
\newblock Efficient memory management for large language model serving with pagedattention.
\newblock In \emph{Proceedings of the 29th symposium on operating systems principles}, pages 611--626.

\bibitem[{Lee et~al.(2024)Lee, Padhi, Ramamurthy, Miehling, Dognin, Nagireddy, and Dhurandhar}]{lee2024programming}
Bruce~W Lee, Inkit Padhi, Karthikeyan~Natesan Ramamurthy, Erik Miehling, Pierre Dognin, Manish Nagireddy, and Amit Dhurandhar. 2024.
\newblock Programming refusal with conditional activation steering.
\newblock \emph{arXiv preprint arXiv:2409.05907}.

\bibitem[{Li et~al.(2026{\natexlab{a}})Li, Huang, Li, Jiang, Wang, Xu, Zhang, Hong, Huang, Xue et~al.}]{li2026perceive}
Hongxing Li, Xiufeng Huang, Dingming Li, Wenjing Jiang, Zixuan Wang, Haolei Xu, Hanrong Zhang, Haiwen Hong, Longtao Huang, Hui Xue, and 1 others. 2026{\natexlab{a}}.
\newblock Perceive-to-reason: Decoupling perception and reasoning for fine-grained visual reasoning.
\newblock \emph{arXiv preprint arXiv:2607.01191}.

\bibitem[{Li et~al.(2026{\natexlab{b}})Li, Bai, Chen, Li, Yang, Lin, and Zhang}]{li2026dynamics}
Zirui Li, Xuefeng Bai, Kehai Chen, Yizhi Li, Jian Yang, Chenghua Lin, and Min Zhang. 2026{\natexlab{b}}.
\newblock Dynamics within latent chain-of-thought: An empirical study of causal structure.
\newblock \emph{arXiv preprint arXiv:2602.08783}.

\bibitem[{Lieberum et~al.(2024)Lieberum, Rajamanoharan, Conmy, Smith, Sonnerat, Varma, Kramár, Dragan, Shah, and Nanda}]{lieberum2024gemmascopeopensparse}
Tom Lieberum, Senthooran Rajamanoharan, Arthur Conmy, Lewis Smith, Nicolas Sonnerat, Vikrant Varma, János Kramár, Anca Dragan, Rohin Shah, and Neel Nanda. 2024.
\newblock \href {https://arxiv.org/abs/2408.05147} {Gemma scope: Open sparse autoencoders everywhere all at once on gemma 2}.
\newblock \emph{Preprint}, arXiv:2408.05147.

\bibitem[{Lightman et~al.(2023)Lightman, Kosaraju, Burda, Edwards, Baker, Lee, Leike, Schulman, Sutskever, and Cobbe}]{lightman2023let}
Hunter Lightman, Vineet Kosaraju, Yuri Burda, Harrison Edwards, Bowen Baker, Teddy Lee, Jan Leike, John Schulman, Ilya Sutskever, and Karl Cobbe. 2023.
\newblock Let's verify step by step.
\newblock In \emph{The Twelfth International Conference on Learning Representations}.

\bibitem[{Lin(2023)}]{neuronpedia}
Johnny Lin. 2023.
\newblock \href {https://www.neuronpedia.org} {Neuronpedia: Interactive reference and tooling for analyzing neural networks}.
\newblock Software available from neuronpedia.org.

\bibitem[{Lin et~al.(2021)Lin, Hilton, and Evans}]{lin2021truthfulqa}
Stephanie Lin, Jacob Hilton, and Owain Evans. 2021.
\newblock Truthfulqa: Measuring how models mimic human falsehoods.
\newblock \emph{arXiv preprint arXiv:2109.07958}.

\bibitem[{Lin et~al.(2025)Lin, Fu, Chen, Chen, Xie, Wang, Cai, Wang, and Ye}]{lin2025controlling}
Zhengkai Lin, Zhihang Fu, Ze~Chen, Chao Chen, Liang Xie, Wenxiao Wang, Deng Cai, Zheng Wang, and Jieping Ye. 2025.
\newblock Controlling thinking speed in reasoning models.
\newblock \emph{arXiv preprint arXiv:2507.03704}.

\bibitem[{Liu et~al.(2025)Liu, Chen, Lu, Ye, Chen, Xing, and Zou}]{liu2025fractional}
Sheng Liu, Tianlang Chen, Pan Lu, Haotian Ye, Yizheng Chen, Lei Xing, and James Zou. 2025.
\newblock Fractional reasoning via latent steering vectors improves inference time compute.
\newblock \emph{arXiv preprint arXiv:2506.15882}.

\bibitem[{Meng et~al.(2022{\natexlab{a}})Meng, Bau, Andonian, and Belinkov}]{meng2022locating}
Kevin Meng, David Bau, Alex Andonian, and Yonatan Belinkov. 2022{\natexlab{a}}.
\newblock Locating and editing factual associations in gpt.
\newblock \emph{Advances in neural information processing systems}, 35:17359--17372.

\bibitem[{Meng et~al.(2022{\natexlab{b}})Meng, Sharma, Andonian, Belinkov, and Bau}]{meng2022mass}
Kevin Meng, Arnab~Sen Sharma, Alex Andonian, Yonatan Belinkov, and David Bau. 2022{\natexlab{b}}.
\newblock Mass-editing memory in a transformer.
\newblock \emph{arXiv preprint arXiv:2210.07229}.

\bibitem[{Miehling et~al.(2026)Miehling, Ramamurthy, Venkateswaran, Ko, Dognin, Singh, Pedapati, Balakrishnan, Riemer, Wei et~al.}]{miehling2026ai}
Erik Miehling, Karthikeyan~Natesan Ramamurthy, Praveen Venkateswaran, Ching-Yun Ko, Pierre Dognin, Moninder Singh, Tejaswini Pedapati, Avinash Balakrishnan, Matthew Riemer, Dennis Wei, and 1 others. 2026.
\newblock Ai steerability 360: A toolkit for steering large language models.
\newblock In \emph{Proceedings of the 64th Annual Meeting of the Association for Computational Linguistics (Volume 3: System Demonstrations)}, pages 436--444.

\bibitem[{Mikolov et~al.(2013)Mikolov, Yih, and Zweig}]{mikolov2013linguistic}
Tom{\'a}{\v{s}} Mikolov, Wen-tau Yih, and Geoffrey Zweig. 2013.
\newblock Linguistic regularities in continuous space word representations.
\newblock In \emph{Proceedings of the 2013 conference of the north american chapter of the association for computational linguistics: Human language technologies}, pages 746--751.

\bibitem[{Olson et~al.(2024)Olson, Ratzlaff, Hinck, Tseng, and Lal}]{olson2024steering}
Matthew~Lyle Olson, Neale Ratzlaff, Musashi Hinck, Shao-yen Tseng, and Vasudev Lal. 2024.
\newblock Steering large language models to evaluate and amplify creativity.
\newblock \emph{arXiv preprint arXiv:2412.06060}.

\bibitem[{Park et~al.(2023)Park, Choe, and Veitch}]{park2023linear}
Kiho Park, Yo~Joong Choe, and Victor Veitch. 2023.
\newblock The linear representation hypothesis and the geometry of large language models.
\newblock \emph{arXiv preprint arXiv:2311.03658}.

\bibitem[{Park et~al.(2025)Park, Du, Yeh, Wang, and Li}]{park2025steer}
Seongheon Park, Xuefeng Du, Min-Hsuan Yeh, Haobo Wang, and Yixuan Li. 2025.
\newblock Steer llm latents for hallucination detection.
\newblock \emph{arXiv preprint arXiv:2503.01917}.

\bibitem[{Sahoo et~al.(2024)Sahoo, Singh, Saha, Jain, Mondal, and Chadha}]{sahoo2024systematic}
Pranab Sahoo, Ayush~Kumar Singh, Sriparna Saha, Vinija Jain, Samrat Mondal, and Aman Chadha. 2024.
\newblock A systematic survey of prompt engineering in large language models: Techniques and applications.
\newblock \emph{arXiv preprint arXiv:2402.07927}.

\bibitem[{Scialanga et~al.(2025)Scialanga, Laugel, Grari, and Detyniecki}]{scialanga2025sake}
Marco Scialanga, Thibault Laugel, Vincent Grari, and Marcin Detyniecki. 2025.
\newblock Sake: Steering activations for knowledge editing.
\newblock \emph{arXiv preprint arXiv:2503.01751}.

\bibitem[{Sellam et~al.(2020)Sellam, Das, and Parikh}]{sellam2020bleurt}
Thibault Sellam, Dipanjan Das, and Ankur~P Parikh. 2020.
\newblock Bleurt: Learning robust metrics for text generation.
\newblock \emph{arXiv preprint arXiv:2004.04696}.

\bibitem[{Seyito{\u{g}}lu et~al.(2024)Seyito{\u{g}}lu, Kuvshinov, Schwinn, and G{\"u}nnemann}]{seyitouglu2024extracting}
Atakan Seyito{\u{g}}lu, Aleksei Kuvshinov, Leo Schwinn, and Stephan G{\"u}nnemann. 2024.
\newblock Extracting unlearned information from llms with activation steering.
\newblock \emph{arXiv preprint arXiv:2411.02631}.

\bibitem[{Turner et~al.(2023)Turner, Thiergart, Leech, Udell, Vazquez, Mini, and MacDiarmid}]{turner2023steering}
Alexander~Matt Turner, Lisa Thiergart, Gavin Leech, David Udell, Juan~J Vazquez, Ulisse Mini, and Monte MacDiarmid. 2023.
\newblock Steering language models with activation engineering.
\newblock \emph{arXiv preprint arXiv:2308.10248}.

\bibitem[{Vogel(2024)}]{vogel2024repeng}
Theia Vogel. 2024.
\newblock \href {https://github.com/vgel/repeng/} {repeng}.

\bibitem[{Wu et~al.(2025)Wu, Ding, Liu, Xia, Huang, Sui, Liu, Wu, Wang, and Tan}]{wu2025sharp}
Junfei Wu, Yue Ding, Guofan Liu, Tianze Xia, Ziyue Huang, Dianbo Sui, Qiang Liu, Shu Wu, Liang Wang, and Tieniu Tan. 2025.
\newblock Sharp: Steering hallucination in lvlms via representation engineering.
\newblock In \emph{Proceedings of the 2025 Conference on Empirical Methods in Natural Language Processing}, pages 14357--14372.

\bibitem[{Wu et~al.(2024)Wu, Arora, Wang, Geiger, Jurafsky, Manning, and Potts}]{wu2024reft}
Zhengxuan Wu, Aryaman Arora, Zheng Wang, Atticus Geiger, Dan Jurafsky, Christopher~D Manning, and Christopher Potts. 2024.
\newblock Reft: Representation finetuning for language models.
\newblock \emph{Advances in Neural Information Processing Systems}, 37:63908--63962.

\bibitem[{Xu et~al.(2026)Xu, Hong, Li, Zhou, Zhang, Huang, Xue, Shen, Lu, and Zhuang}]{xu2026seeing}
Haolei Xu, Haiwen Hong, Hongxing Li, Rui Zhou, Yang Zhang, Longtao Huang, Hui Xue, Yongliang Shen, Weiming Lu, and Yueting Zhuang. 2026.
\newblock Seeing but not thinking: Routing distraction in multimodal mixture-of-experts.
\newblock In \emph{Proceedings of the 64th Annual Meeting of the Association for Computational Linguistics (Volume 1: Long Papers)}, pages 31164--31178.

\bibitem[{Xu et~al.(2025{\natexlab{a}})Xu, Yan, Shen, Zhang, Hou, Jiang, Song, Lu, Xiao, and Zhuang}]{xu2025mind}
Haolei Xu, Yuchen Yan, Yongliang Shen, Wenqi Zhang, Guiyang Hou, Shengpei Jiang, Kaitao Song, Weiming Lu, Jun Xiao, and Yueting Zhuang. 2025{\natexlab{a}}.
\newblock Mind the gap: Bridging thought leap for improved chain-of-thought tuning.
\newblock \emph{arXiv preprint arXiv:2505.14684}.

\bibitem[{Xu et~al.(2025{\natexlab{b}})Xu, Wang, Xu, Xu, Wang, Deng, Yao, Zheng, Chen, and Zhang}]{xu2025easyedit2}
Ziwen Xu, Shuxun Wang, Kewei Xu, Haoming Xu, Mengru Wang, Xinle Deng, Yunzhi Yao, Guozhou Zheng, Huajun Chen, and Ningyu Zhang. 2025{\natexlab{b}}.
\newblock Easyedit2: An easy-to-use steering framework for editing large language models.
\newblock \emph{arXiv preprint arXiv:2504.15133}.

\bibitem[{Yang et~al.(2024)Yang, Yang, Zhang, Hui, Zheng, Yu, Li, Liu, Huang, Wei, Lin, Yang, Tu, Zhang, Yang, Yang, Zhou, Lin, Dang, Lu, Bao, Yang, Yu, Li, Xue, Zhang, Zhu, Men, Lin, Li, Xia, Ren, Ren, Fan, Su, Zhang, Wan, Liu, Cui, Zhang, and Qiu}]{qwen2025qwen25technicalreport}
An~Yang, Baosong Yang, Beichen Zhang, Binyuan Hui, Bo~Zheng, Bowen Yu, Chengyuan Li, Dayiheng Liu, Fei Huang, Haoran Wei, Huan Lin, Jian Yang, Jianhong Tu, Jianwei Zhang, Jianxin Yang, Jiaxi Yang, Jingren Zhou, Junyang Lin, Kai Dang, and 22 others. 2024.
\newblock Qwen2.5 technical report.
\newblock \emph{arXiv preprint arXiv:2412.15115}.

\bibitem[{Yao et~al.(2024)Yao, Duan, Xu, Cai, Sun, and Zhang}]{yao2024survey}
Yifan Yao, Jinhao Duan, Kaidi Xu, Yuanfang Cai, Zhibo Sun, and Yue Zhang. 2024.
\newblock A survey on large language model (llm) security and privacy: The good, the bad, and the ugly.
\newblock \emph{High-Confidence Computing}, 4(2):100211.

\bibitem[{Zhang et~al.(2026{\natexlab{a}})Zhang, Zhang, Wang, Su, Wang, Wang, Yuan, Nie, Duan, Xue et~al.}]{zhang2026locate}
Hengyuan Zhang, Zhihao Zhang, Mingyang Wang, Zunhai Su, Yiwei Wang, Qianli Wang, Shuzhou Yuan, Ercong Nie, Xufeng Duan, Qibo Xue, and 1 others. 2026{\natexlab{a}}.
\newblock Locate, steer, and improve: A practical survey of actionable mechanistic interpretability in large language models.
\newblock \emph{arXiv preprint arXiv:2601.14004}.

\bibitem[{Zhang et~al.(2026{\natexlab{b}})Zhang, Yao, Qin, Xu, Zhu, Yu, Wang, Tang, Gu, Deng et~al.}]{zhang2026towards}
Ningyu Zhang, Yunzhi Yao, Jiaxin Qin, Haoming Xu, Yuqi Zhu, Zeping Yu, Mengru Wang, Yuqi Tang, Jia-Chen Gu, Shumin Deng, and 1 others. 2026{\natexlab{b}}.
\newblock Towards principled knowledge editing methods for large language model reasoning.
\newblock \emph{Nature Machine Intelligence}, pages 1--12.

\bibitem[{Zhang et~al.(2024)Zhang, Yao, Tian, Wang, Deng, Wang, Xi, Mao, Zhang, Ni et~al.}]{zhang2024comprehensive}
Ningyu Zhang, Yunzhi Yao, Bozhong Tian, Peng Wang, Shumin Deng, Mengru Wang, Zekun Xi, Shengyu Mao, Jintian Zhang, Yuansheng Ni, and 1 others. 2024.
\newblock A comprehensive study of knowledge editing for large language models.
\newblock \emph{arXiv preprint arXiv:2401.01286}.

\bibitem[{Zhang et~al.(2026{\natexlab{c}})Zhang, Ma, Hou, Bai, Chen, Xiang, Yu, and Zhang}]{zhang2026evaluating}
Yu~Zhang, Jinlong Ma, Yongshuai Hou, Xuefeng Bai, Kehai Chen, Yang Xiang, Jun Yu, and Min Zhang. 2026{\natexlab{c}}.
\newblock Evaluating and steering modality preferences in multi-modal llms.
\newblock In \emph{Forty-third International Conference on Machine Learning}.

\bibitem[{Zhao et~al.(2023)Zhao, Zhou, Li, Tang, Wang, Hou, Min, Zhang, Zhang, Dong et~al.}]{zhao2023survey}
Wayne~Xin Zhao, Kun Zhou, Junyi Li, Tianyi Tang, Xiaolei Wang, Yupeng Hou, Yingqian Min, Beichen Zhang, Junjie Zhang, Zican Dong, and 1 others. 2023.
\newblock A survey of large language models.
\newblock \emph{arXiv preprint arXiv:2303.18223}, 1(2).

\end{thebibliography}
